\documentclass[sigconf]{acmart}
\graphicspath{ {images/} }
\usepackage{booktabs}
\usepackage[labelformat=simple]{subcaption}

\copyrightyear{2020}
\acmYear{2020}
\acmConference[GECCO '20]{Genetic and Evolutionary Computation Conference}{July 8--12, 2020}{Cancún, Mexico}
\acmBooktitle{Genetic and Evolutionary Computation Conference (GECCO '20), July 8--12, 2020, Cancún, Mexico}
\acmDOI{10.1145/3377930.3389824}

\setcopyright{none}
\renewcommand\footnotetextcopyrightpermission[1]{}
\begin{document}
\title{Exploring the Evolution of GANs through Quality Diversity}

\author{Victor Costa}
\affiliation{%
	\institution{University of Coimbra, CISUC, DEI}
	\city{Coimbra}
	\country{Portugal}
}
\email{vfc@dei.uc.pt}

\author{Nuno Louren\c{c}o}
\affiliation{%
	\institution{University of Coimbra, CISUC, DEI}
	\city{Coimbra}
	\country{Portugal}
}
\email{naml@dei.uc.pt}

\author{Jo\~{a}o Correia}
\affiliation{%
	\institution{University of Coimbra, CISUC, DEI}
	\city{Coimbra}
	\country{Portugal}
}
\email{jncor@dei.uc.pt}

\author{Penousal Machado}
\affiliation{%
	\institution{University of Coimbra, CISUC, DEI}
	\city{Coimbra}
	\country{Portugal}
}
\email{machado@dei.uc.pt}

\begin{abstract}
Generative adversarial networks (GANs) achieved relevant advances in the field of generative algorithms, presenting high-quality results mainly in the context of images. However, GANs are hard to train, and several aspects of the model should be previously designed by hand to ensure training success. In this context, evolutionary algorithms such as COEGAN were proposed to solve the challenges in GAN training. Nevertheless, the lack of diversity and premature optimization can be found in some of these solutions. We propose in this paper the application of a quality-diversity algorithm in the evolution of GANs. The solution is based on the Novelty Search with Local Competition (NSLC) algorithm, adapting the concepts used in COEGAN to this new proposal. We compare our proposal with the original COEGAN model and with an alternative version using a global competition approach. The experimental results evidenced that our proposal increases the diversity of the discovered solutions and leverage the performance of the models found by the algorithm. Furthermore, the global competition approach was able to consistently find better models for GANs.
\end{abstract}



\maketitle

\section{Introduction}
Generative Adversarial Networks (GANs) \cite{NIPS2014_5423} are an adversarial model that gained a lot of relevance in recent years, mainly by the success in generative tasks.
Even though GANs can be applied in several contexts such as image, video, sound, and text, the GAN model is popular for the impressive results concerning the quality of created samples in the context of images.
The GAN model consists of two neural networks: one generator and one discriminator, trained in an adversarial way.
A successful GAN training produces strong generative and discriminative components.

Despite the compelling results, the training of GANs is challenging and is frequently affected by the presence of problems such as the vanishing gradient and the mode collapse~\cite{brock2018large,fedus2017many}.
Improvements over the original model were introduced to handle these issues, but they are still a problem~\cite{salimans2016improved,gulrajani2017improved,arjovsky2017wasserstein}.
Alternatives loss functions such as in WGAN~\cite{arjovsky2017wasserstein}, LSGAN~\cite{mao2017least}, and SN-GAN~\cite{miyato2018spectral} were proposed to improve the model.
Besides, architectural guides and strategies were developed to minimize these issues~\cite{radford2015unsupervised, karras2018progressive}.

Another strategy to improve the training of GANs is the application of evolutionary algorithms.
In this context, recent solutions were designed to improve the training process and the quality of the outcome~\cite{al2018towards,costa2019evaluating,costa2019coevolution,garciarena2018evolved,wang2018evolutionary,toutouh2019spatial}.
These proposals incorporate mechanisms such as neuroevolution, coevolution, and spatial coevolution, making use of evolutionary pressure to achieve efficient models.

Coevolutionary Generative Adversarial Networks (COEGAN)~\cite{costa2019evaluating,costa2019coevolution} is a solution inspired on NEAT~\cite{neat} and DeepNEAT~\cite{miikkulainen2017evolving} that applies a coevolution model to evolve GANs.
Experimental results show that the model provides a more reliable GAN training when compared to regular GANs in equivalent scenarios.
However, the lack of diversity evidenced in the experimental evaluation affects the quality of the results, leaving space for improvement of the algorithm that drives the evolutionary process.
Thus, we study in this work a mechanism of novelty to be applied in COEGAN in order to improve the exploration of solutions.
Quality Diversity (QD) algorithms are a class of solutions that can be used to enhance the population and produce a diversity of efficient individuals~\cite{pugh2015confronting}.
At the best of our knowledge, no proposed solutions combining evolutionary algorithms and GANs use novelty search or QD mechanisms in their approaches.

In this paper, we propose a new model combining concepts used on COEGAN and a quality diversity algorithm for guiding the evolution of GANs.
Therefore, instead of strategies such as the speciation based on NEAT to support evolution, we propose the use of Novelty Search with Local Competition (NSLC)~\cite{lehman2011evolving}, a quality diversity algorithm that uses mechanisms of novelty in the search for efficient solutions.
We aim to improve the exploration of the search space and achieve better models for generators and discriminators.

To validate our proposal, experiments were conducted using the MNIST~\cite{lecun1998mnist} and CelebA~\cite{liu2015deep} datasets.
We compare the results between the original COEGAN approach and two variations of our proposal: COEGAN with NSLC and with an alternative using a global competition strategy.
The results evidenced that the exploration of solutions was improved with the QD algorithm, leading to the discovery of more efficient models for GANs.
Besides, the global competition version achieved the best results concerning the quality of samples created by generators.

The remainder of this paper is organized as follows:
Section~\ref{sec:background} introduces concepts related to GANs, also describing the works associated with evolutionary algorithms;
Section~\ref{sec:model} presents our approach to use a quality diversity algorithm in the evolution of GANs;
Section~\ref{sec:experiments} displays the experimental results of this approach;
finally, Section~\ref{sec:conclusions} presents our conclusions.

\section{Background and Related Works} \label{sec:background}
Generative Adversarial Networks (GANs) represent an adversarial model composed of two neural networks: a generator and a discriminator.
The discriminator receives a dataset as input and has to distinguish between samples of this dataset and fake samples.
The generator is responsible for producing synthetic data in order to fool the discriminator.
As the training progresses, both the generator and the discriminator improve their tasks, resulting in strong generative and discriminative components at the end of a successful training.

The regular GAN training uses backpropagation and gradient descent in both neural networks.
Thus, the loss function of the discriminator is defined as follows:
\begin{equation}
J^{(D)}(D,G) = -\mathbb{E}_{x \sim p_{data}}[\log D(x)] - \mathbb{E}_{z \sim p_z}[\log(1 - D(G(z)))].
\label{eq:discriminator}
\end{equation}

For the generator, the non-saturating version of the loss function is defined by:
\begin{equation}
J^{(G)}(G) = - \mathbb{E}_{z \sim p_z}[\log(D(G(z)))].
\label{eq:generator}
\end{equation}

In Eq.~\eqref{eq:discriminator}, $p_{data}$ represents the dataset used as input to the discriminator.
In Eq.~\eqref{eq:discriminator} and Eq.~\eqref{eq:generator}, $z$ is the latent space used as input to the generator, $p_z$ is the latent distribution, $G$ is the generator, and $D$ represents the discriminator.

GANs are hard to train and a trial-and-error approach is frequently used to get consistent results.
The equilibrium of forces between the discriminator and the generator is frequently the cause of problems in training.
In the vanishing gradient problem, the discriminator or generator is so powerful that it becomes almost perfect in its task, leading to the stagnation of training progress.
The mode collapse occurs when the generator captures only a small fraction of the input distribution, limiting the diversity of produced samples.

In GANs, the Fr\'{e}chet Inception Distance (FID)~\cite{heusel2017gans} is often used to evaluate the performance of the generators.
The FID metric uses the Inception Net~\cite{szegedy2016rethinking} (trained on ImageNet~\cite{russakovsky2015imagenet}) to transform images into a feature space, which is interpreted as a continuous multivariate Gaussian.
This process is applied to samples from the input dataset and synthetic samples created by the generator.
The mean and covariance of the two resulting Gaussians are estimated and the Fr\'{e}chet distance between them is given by:
\begin{equation}
FID(x,g) = ||\mu_x - \mu_g||_2^2 + Tr(\varSigma_x + \varSigma_g - 2(\varSigma_x\varSigma_g)^{1/2}),
\label{eq:fid}
\end{equation}
with $\mu_x$, $\varSigma_x$, $\mu_g$, and $\varSigma_g$ representing the mean and covariance estimated for the input dataset $x$ and fake samples $g$, respectively.
This metric is capable of quantifying the quality and diversity of the generative model.

The use of evolutionary algorithms to train and evolve GANs was recently proposed~\cite{al2018towards,costa2019evaluating,costa2019coevolution,garciarena2018evolved,wang2018evolutionary,toutouh2019spatial}.
The solutions present a diverse set of strategies to not only overcome common GAN problems but also to provide better quality on the produced samples.

E-GAN~\cite{wang2018evolutionary} evolve GANs using a variation operator that switches the loss function of the generator through generations.
In this case, a single-fixed discriminator is used as the adversarial for the population of generators, with the former using a fixed architecture and loss function, and the latter varying only the loss function.
The architectures of the generator and the discriminator are based on DCGAN~\cite{radford2015unsupervised}.
In~\cite{garciarena2018evolved}, Pareto set approximation was used in a neuroevolution algorithm to evolve GANs.
In this case, the architectures of the networks are dynamic and change according to the variation operators.
Lipizzaner~\cite{al2018towards} uses spatial coevolution to train GANs.
However, the networks of the discriminator and generator are fixed and only the internal parameters (e.g., weights) change through evolution.
A further improvement over Lipizzaner, called Mustangs~\cite{toutouh2019spatial}, applies the E-GAN dynamic loss function to the algorithm while keeping the same spatial coevolution strategy of Lipizzaner.

COEGAN~\cite{costa2019evaluating,costa2019coevolution} was inspired by NEAT~\cite{neat} and DeepNEAT~\cite{miikkulainen2017evolving} to design an evolutionary algorithm for GANs, using mechanisms such as speciation to protect innovation during the evolution.
In COEGAN, the fitness used for generators is based on the FID score (Eq.~\eqref{eq:fid}).
The use of FID was designed to put selection pressure in generators and guide the evolution of the population in producing better samples.
For discriminators, fitness is based on the loss function represented by Eq.~\eqref{eq:discriminator}.

Nondominated Sorting Genetic Algorithm II (NSGA-II)~\cite{deb2002fast} is a well-known solution in the class of Multi-Objective Evolutionary Algorithms (MOEAs) that uses an elitist method to implement a Pareto-based search approach.
In NSGA-II, an algorithm to sort solutions and determine nondominated fronts is used on the generation of the next populations.
Besides, a crowding-distance computation is used as a second criterion to prioritize solutions in less explored spaces.
This algorithm also uses an archive to keep the previously explored solutions and improve diversity.
Individuals are usually inserted into the archive using a probabilistic approach.

Quality Diversity (QD) algorithms are a family of evolutionary algorithms aiming to find a diversity of viable solutions for the target problem~\cite{pugh2015confronting}.
Novelty Search with Local Competition (NSLC)~\cite{lehman2011evolving} uses a Pareto-based MOEA, such as NSGA-II, to promote the quality and diversity of solutions.
NSLC uses as objectives the quality and novelty of individuals according to a local neighborhood.
Therefore, when combined with NSGA-II, NSLC does not use the crowding-distance mechanism because the novelty criterion already produces the desired diversity.

MAP-Elites~\cite{mouret2015illuminating} is another solution in the class of QD algorithms.
In MAP-Elites, an archive of phenotypes is kept in order to explore the diversity of high-performing solutions.
Thus, at each step, an item of the archive is chosen to produce the offspring.
Then, the performance is calculated and the newly generated individual is placed into the archive in the position determined by the feature space, replacing older individuals in case of better performance.
The MAP-Elites algorithm was also considered to be the subject of this research but the cost to explore and maintain an archive of high-dimensional neural networks was considered too high.

\section{Methods} \label{sec:model}
We describe in this section our proposal to use a Quality Diversity (QD) algorithm to evolve GANs.
This new model is based on the original COEGAN proposal~\cite{costa2019coevolution,costa2019evaluating}, adapted to be guided by a different evolutionary algorithm.
Thus, first we introduce the fundamentals of the COEGAN model.
Then, we describe our approach to applying the QD algorithm with COEGAN.

\subsection{COEGAN} \label{sec:coegan}
In COEGAN, the genome consists of an array of genes that are directly mapped into sequential layers of a neural network, forming the phenotype of individuals.
This approach was inspired by NEAT~\cite{neat} and its extension DeepNEAT~\cite{miikkulainen2017evolving}.
The genes describe either a linear, convolution or transpose convolution layer (also known as deconvolution layer), depending on the type of individuals.
Generators allow linear and transpose convolution layers, and discriminators allow linear and convolution layers.
In addition, each gene holds internal parameters specific to the type of layer.
Parameters such as the activation function, the number of output features, and the number of output channels are subject to variation operators.

The variation operators are based on mutation, with the possibilities to add, remove or change genes.
The addition operator inserts a new gene into the genome.
This new layer is randomly drawn from a set of possible layers: linear and convolution for discriminators; linear and transpose convolution for generators.
The removal operation randomly chooses an existing gene and excludes it from the genome.
The change operation modifies the internal attributes such as the activation function of an existing layer.
In this case, the activation function is randomly chosen from the set: ReLU, LeakyReLU, ELU, Sigmoid, and Tanh.
For dense and convolution layers, the number of output features and channels can also be mutated, respectively.
The mutation of these attributes follows a uniform distribution, delimited by a predefined range.

In COEGAN, the weights of parents are transferred to the children whenever it is possible~\cite{costa2019coevolution,costa2019evaluating}.
This mechanism of transference ensures that the information achieved on training in previous generations is kept during the whole evolution process.
Therefore, not only the final models achieved by COEGAN are important but also the entire process used in the discovery of them.

Competitive coevolution was used to model the COEGAN algorithm, creating two independent subpopulations of generators and discriminators.
Thus, the evaluation phase matches generators and discriminators for training and evaluation.
The \textit{all vs. all} approach is used to pair each generator with each discriminator in the regular GAN training algorithm.

A fitness sharing strategy is used to protect the species and promote innovation.
Therefore, each subpopulation of generators and discriminators is divided into species.
The speciation mechanism was inspired by NEAT and aims to promote innovation in the population, ensuring that recently modified individuals will be trained for enough generations to be comparable to older individuals with respect to the fitness value.
A similarity function comparing pairs of genomes is used to group individuals into species.

At the evaluation phase, the fitness used for discriminators is based on the loss function of the discriminator in the original GAN model, represented by Eq.~\eqref{eq:discriminator}.
For generators, the fitness is the FID score, represented by Eq.~\eqref{eq:fid}.

\subsection{Quality Diversity in COEGAN}
In this work, we propose a new evolutionary algorithm to guide the COEGAN training.
Therefore, we replace the NEAT-based evolutionary algorithm used in COEGAN with an approach based on Novelty Search with Local Competition (NSLC)~\cite{lehman2011evolving}.
As originally proposed in~\cite{lehman2011evolving}, we use Nondominated Sorting Genetic Algorithm II (NSGA-II)~\cite{deb2002fast} as the Multi-Objective Evolutionary Algorithm (MOEA) for NSLC.

We kept some aspects used in COEGAN regarding the genotype representation.
Thus, in our proposal, the genome and variation operators are the same originally used in COEGAN, described in Section~\ref{sec:coegan}.
The competitive coevolution model is also the same used in COEGAN.
Following we describe the differences and new aspects of the model proposed in this work.

We designed the pairing between individuals at the evaluation phase according to the \textit{all vs. all} coevolution model used in COEGAN.
However, as NSGA-II uses an elitist approach to select individuals to form the next generation, both the current population and the derived offspring should be evaluated.
For this, we match generators from the current population with discriminators from the offspring, and discriminators from the current population with generators from the offspring population.
This ensures the progress of all individuals through generations, making it possible to properly select them when targeting for quality and diversity.
The main drawback here is that now the double amount of individuals should be evaluated, increasing the execution cost of the algorithm.
However, it is important to note that we keep the same number of training steps for each individual when compared to the original COEGAN approach.

Tournament is applied to select individuals for reproduction.
As proposed by the NSGA-II algorithm, a dominance operator is used to determine the result of the tournament between a set of individuals.
In this research, we use the constrained version of the operator, ensuring that the population does not deviate too much from the objective defined by the fitness function.
Thus, we use not only the concept of dominance but also the feasibility of solutions.

NSGA-II designs the dominance operator using the ranking of solutions determined by the nondominated sorting algorithm, aiming to obtain solutions in the Pareto-optimal front.
The feasibility concept ensures that, when comparing two solutions $s_i$ and $s_j$, the fitness function meets the constraint $f(s_j) < 2f(s_i)$, otherwise $s_j$ is considered unfeasible.
In summary, the solution $s_i$ constrained-dominates a solution $s_j$ when $s_i$ is feasible and $s_j$ is not, or both solutions are feasible and $s_i$ dominates $s_j$.
Note that one of the solutions will always be feasible, i.e., the case that both solutions are unfeasible is not possible.

The definition of the neighborhood is paramount to the NSLC algorithm, being used in both the quality and diversity criteria.
In order to determine the neighbors of each individual, we use the distance between the architectures of the neural networks of individuals, which is directly defined by the similarity between genomes.
This distance is the same used originally in COEGAN to group individuals into species.
Two individuals are considered equal if they have the same genome, i.e., the same sequence of genes, disregarding other characteristics like age or the number of samples currently used in the training.
It is important to note that the neighborhood calculation considers not only the current population but also the archive of previous solutions.
This archive is filled following a probabilistic approach, i.e., at each generation individuals are inserted into the archive with a predefined probability.

In NSLC, $n$ nearest neighbors of an individual are selected to calculate the innovation and the competition objectives.
Innovation is defined by the average distance between the individual and the neighborhood.
The competition objective is defined by the number of neighbors the individual outperforms with respect to the fitness value.
In our proposal, the fitness is the same used in COEGAN: Eq.~\eqref{eq:discriminator} for discriminators and Eq.~\eqref{eq:generator} for generators.

The innovation criteria make it possible to better explore the available architectures characterized by the genotype representation.
When combined with the strategy used to calculate the competition score, different niches can be efficiently explored to leverage the search space.
Compared to the original COEGAN approach, we expect to improve the diversity of solutions and eventually find better results concerning the FID score.
In COEGAN, the number of species is fixed and previously defined, being a limitation over the capacity of innovation for individuals that need to survive through generations to show consistent performances.
In COEGAN guided by NSLC, the exploration of the search space is improved by the novelty criterion, using the quality definition to guide the population through the objective of obtaining better solutions.

\section{Experiments} \label{sec:experiments}
To validate our proposal, we present an experimental analysis of the application of Quality Diversity in the evolution of GANs~\footnote{Code available at https://github.com/vfcosta/qd-coegan.}.
Therefore, we conduct experiments using MNIST in order to evidence the performance of the algorithm proposed in this work compared with the original COEGAN model.
We also design experiments with an alternative version of the solution which uses a global competition mechanism, i.e., the neighborhood is not limited by a constant $n$ and uses all individuals available.
We call this version of the algorithm Novelty Search with Global Competition (NSGC), inspired by the global competition approach experimented in~\cite{lehman2011evolving}.
Therefore, we refer to the original COEGAN proposal, COEGAN trained with the NSLC algorithm and trained with NSGC as COEGAN, COEGAN+NSLC, and COEGAN+NSGC, respectively.

The FID score was used to measure the quality of the produced samples.
Besides, the strategy proposed in~\cite{zhang2018stackgan++} was applied to present the visual distribution of image samples, using t-SNE~\cite{maaten2008visualizing} to embed samples into a two-dimensional space.
Further experiments with the CelebA dataset were made to compare our method with a non-evolutionary GAN approach in a more complex dataset.

\subsection{Experimental Setup}
Table \ref{table:setup} lists the parameters used in our experiments.
These parameters were selected based on preliminary experiments and the experiments presented in~\cite{costa2019coevolution,costa2019evaluating}.
The number of generations used in all experiments is $50$.
Each population of generators and discriminators contains $10$ individuals.
We use a probability of 30\%, 10\% and 10\% for the add, remove and change mutations, respectively.
The genome of generators and discriminators was limited to four genes, representing a network of four layers in the maximum allowed setup.
This setup is sufficient to discover efficient solutions for the experiments with the MNIST dataset.

For the original COEGAN, we use $3$ species in each population of generators and discriminators.
For COEGAN with NSLC, the number of neighbors $n$ is limited to $3$ and the probability to insert individuals into the archive is $10\%$.
The global version of the QD algorithm does not limit the neighborhood, using all individuals when calculating the novelty and competition values.

Figures in this section display plots with curves representing the average of the results from $15$ independent executions, with a confidence interval of $95\%$.

\begin{table}
	\caption{Experimental parameters.}
	\begin{center}\begin{tabular}{c|c}
			\textbf{Evolutionary Parameters} & \textbf{Value} \\
			\hline
			Number of generations & 50 \\
			Population size (generators) & 10 \\
			Population size (discriminators) & 10 \\
			Add Layer rate & 30\% \\
			Remove Layer rate & 10\% \\
			Change Layer rate & 10\% \\
			Output channels range & [32, 256] \\
			Tournament $k_t$ & 2 \\
			FID samples & 1024 \\
			Genome Limit & 4 \\
			Species & 3 \\
			Neighborhood size $n$ & 3 \\
			Archive probability & 10\% \\
			\textbf{GAN Parameters} & \textbf{Value} \\
			\hline
			Batch size & 64 \\
			Batches per generation & 50 \\
			Optimizer & Adam \\
			Learning rate & 0.001
	\end{tabular}\end{center}
	\label{table:setup}
\end{table}

\subsection{Results}
We present in this section the results of the experimental analysis, comparing the solutions using the QD algorithm with the previously proposed COEGAN model.
First, we present the results using the MNIST dataset.
Then, we provide a further analysis with CelebA, a more complex dataset, comparing our proposal with a regular GAN approach.

\begin{figure}[h]
	\includegraphics[width=0.4\textwidth]{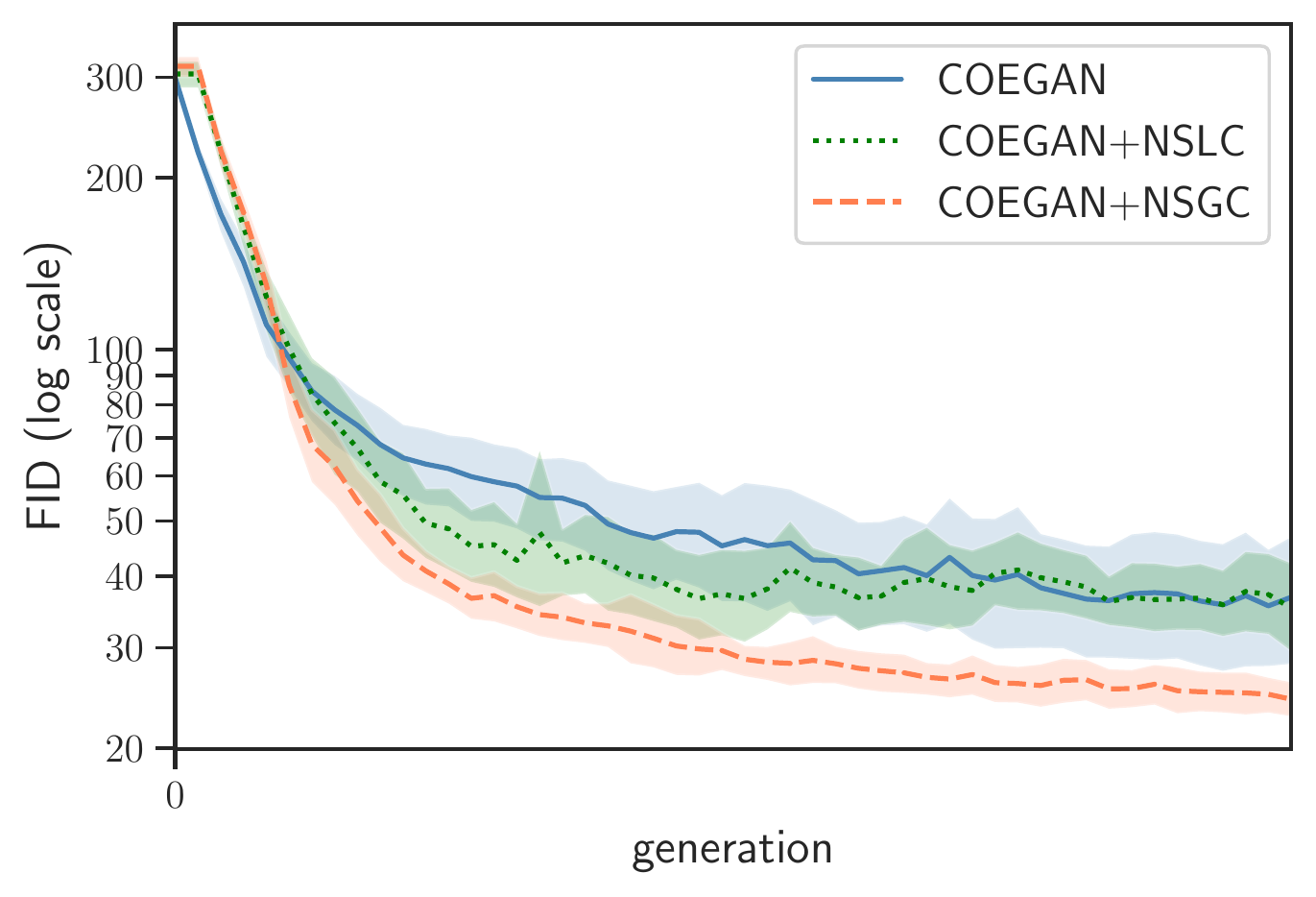}
	\caption{Best FID Score on the MNIST dataset.}
	\label{fig:mnist_fid_score_g_best}
\end{figure}

Figure~\ref{fig:mnist_fid_score_g_best} shows the FID score of the best individuals for each generation when training with the MNIST dataset.
We can see that the COEGAN+NSLC solution outperforms the original COEGAN model by a small margin until half generations but has equivalent performance at the end.
This effect is mostly due to the increased exploration capability given by the QD algorithm, which produces a more diverse population but causes less focused evolution of more fitted individuals.
Besides, COEGAN+NSGC, the global competition variation, has better performance than COEGAN+NSLC and the original COEGAN approach.
The results obtained by comparing the global and local competition versions of the algorithm are similar to results presented in~\cite{deb2002fast}, where the global version also achieved better fitness than NSLC.

\begin{table}
	\caption{Average FID score of best generators after training with the MNIST dataset.}
	\begin{center}\begin{tabular}{c|c}
			\textbf{Algorithm} & \textbf{FID Score} \\
			\hline
			COEGAN & $36.8\pm18.6$ \\
			COEGAN+NSLC & $35.2\pm12.5$ \\
			COEGAN+NSGC & $24.3\pm3.3$
	\end{tabular}\end{center}
	\label{table:fid_mnist}
\end{table}

\begin{figure}[h]
	\includegraphics[width=0.4\textwidth]{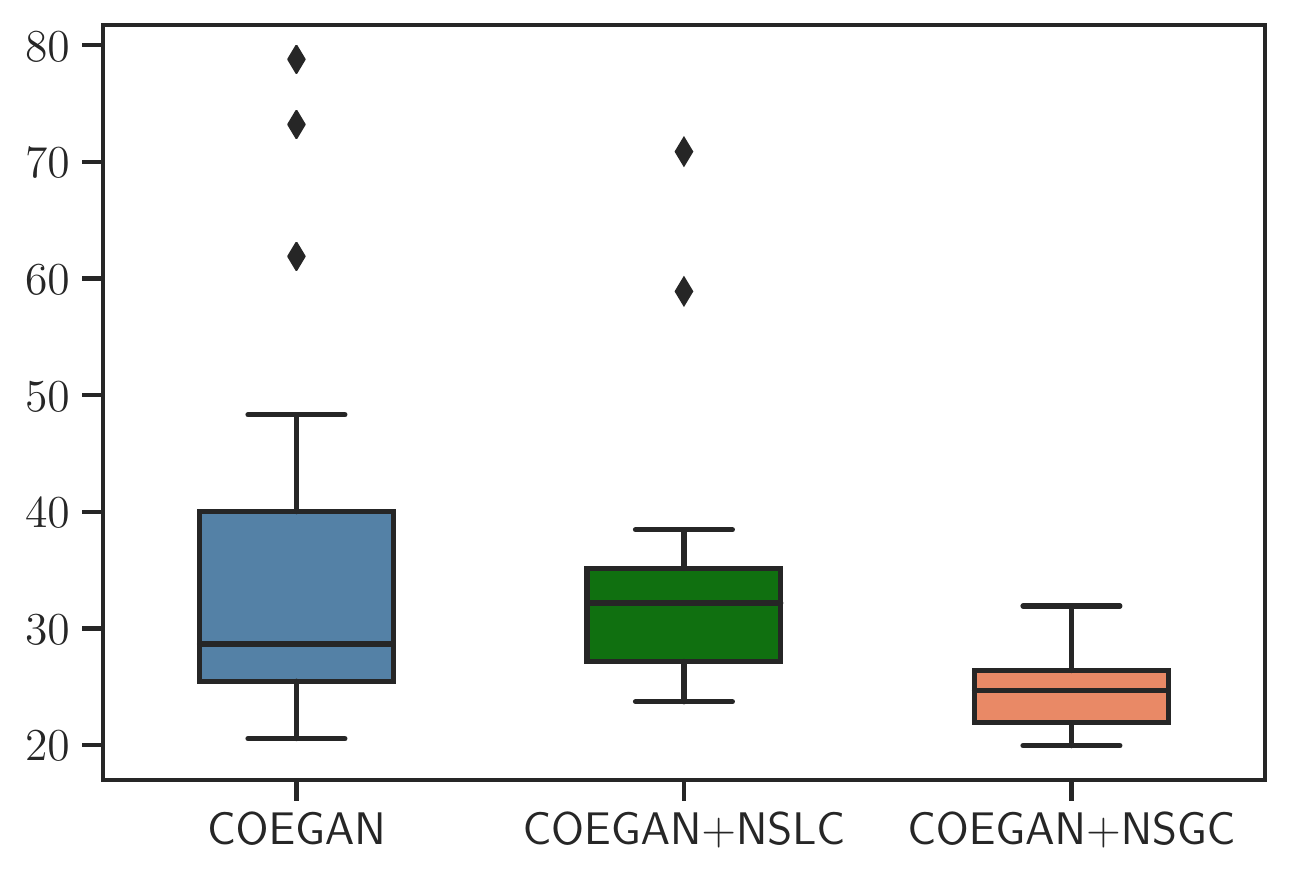}
	\caption{Boxplot of the FID score on MNIST dataset showing the performance of best generators computed for each independent run.}
	\label{fig:fid_mnist_boxplot}
\end{figure}

We can see in Figure~\ref{fig:fid_mnist_boxplot} and Table~\ref{table:fid_mnist} that COEGAN+NSGC consistently achieved better results than the other solutions.
COEGAN provides more unstable results when compared to the global approach, presenting a higher standard deviation in FID values when trained with the experimental setup described in Table~\ref{table:setup}.
This effect is also present in the experiments with COEGAN+NSLC, indicating that the high diversity produced by our experimental setup affects the results with respect to the FID score.
Therefore, we conclude that the diversity provided by COEGAN+NSGC is sufficient to achieve better results in our approach to train GANs.

To support our analysis, we statistically test the significance of our findings.
We assume that the results do not follow a normal distribution, as the normality test (Shapiro-Wilk, $\alpha = 0.05$) rejected this hypothesis for COEGAN and COEGAN+NSLC ($p < 0.001$).
Then, we used a non-parametric test (Mann-Whitney U, $\alpha = 0.05$) to perform a pairwise comparison between the solutions evaluated in this work.
We found that the improvement of COEGAN+NSGC over COEGAN is statistically significant ($p = 0.008$).
Moreover, the performance improvement of COEGAN+NSGC over COEGAN+NSLC is also statistically significant ($p = 0.0001$).
We found no statistical difference between COEGAN and COEGAN+NSLC ($p = 0.17$).
Further experiments should be performed to assess the influence of the experimental setup in these results.

\begin{figure}[h]
	\includegraphics[width=0.4\textwidth]{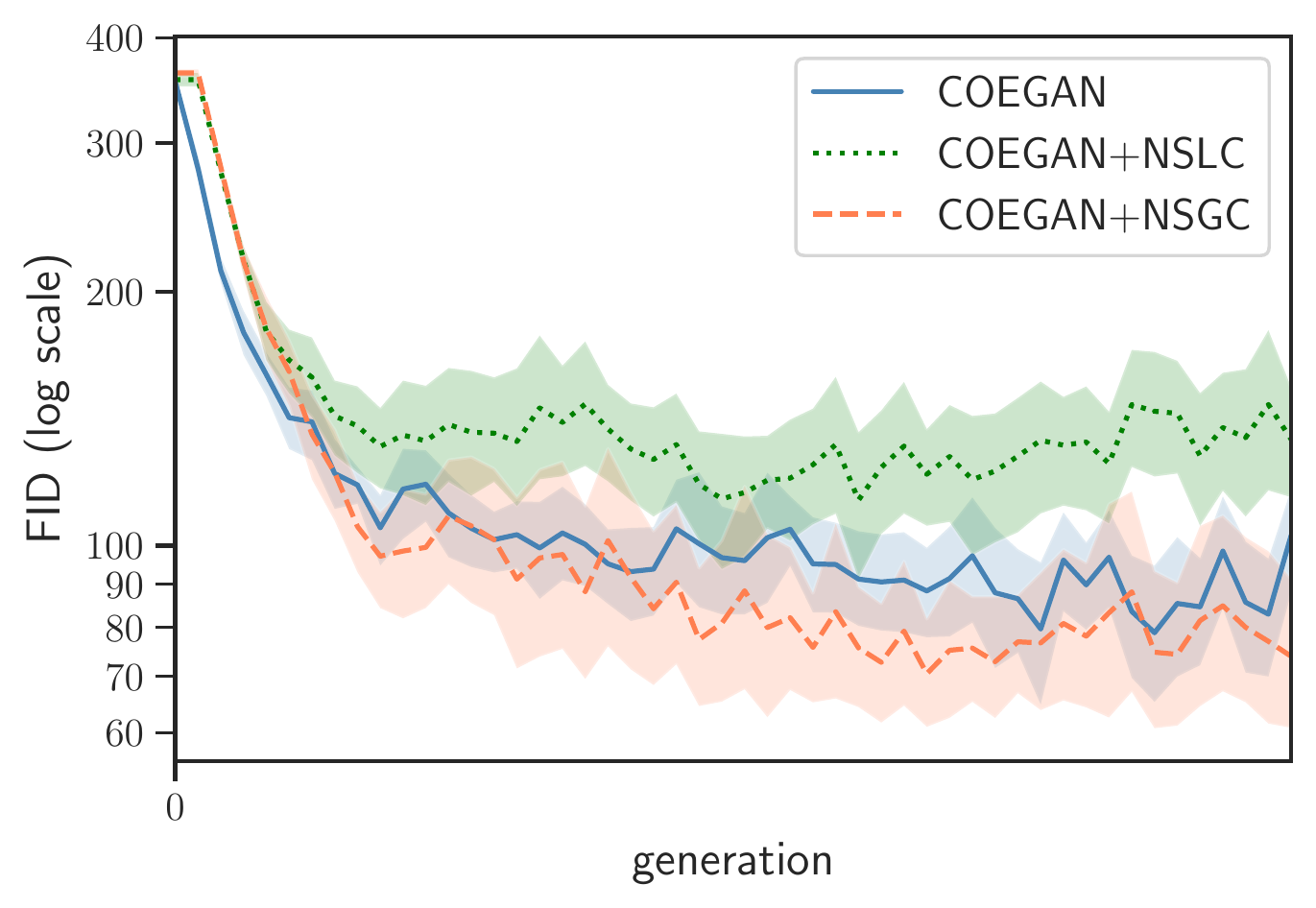}
	\caption{Average FID Score on the MNIST dataset.}
	\label{fig:mnist_fid_score_g_mean}
\end{figure}

We also show in Figure~\ref{fig:mnist_fid_score_g_mean} the FID scores of all individuals in the population of generators when training with MNIST.
This result evidences that COEGAN+NSLC has a better exploration of the search space, increasing the diversity and leading to the discovery of not only good individuals but also less efficient solutions.
The local competition approach used in NSLC has a stronger effect on the protection of innovation, as the competition calculation uses the fitness values from similar individuals, i.e., it uses the $n$ closest neighbors concerning the architectural similarity.
The effect of this is the discovery of niches that are not efficient in terms of fitness.
In the scenario of global competition, individuals have to outperform a broader range of solutions in order to survive through generations, leading to the convergence of better models.

\begin{figure}[h]
	\includegraphics[width=0.4\textwidth]{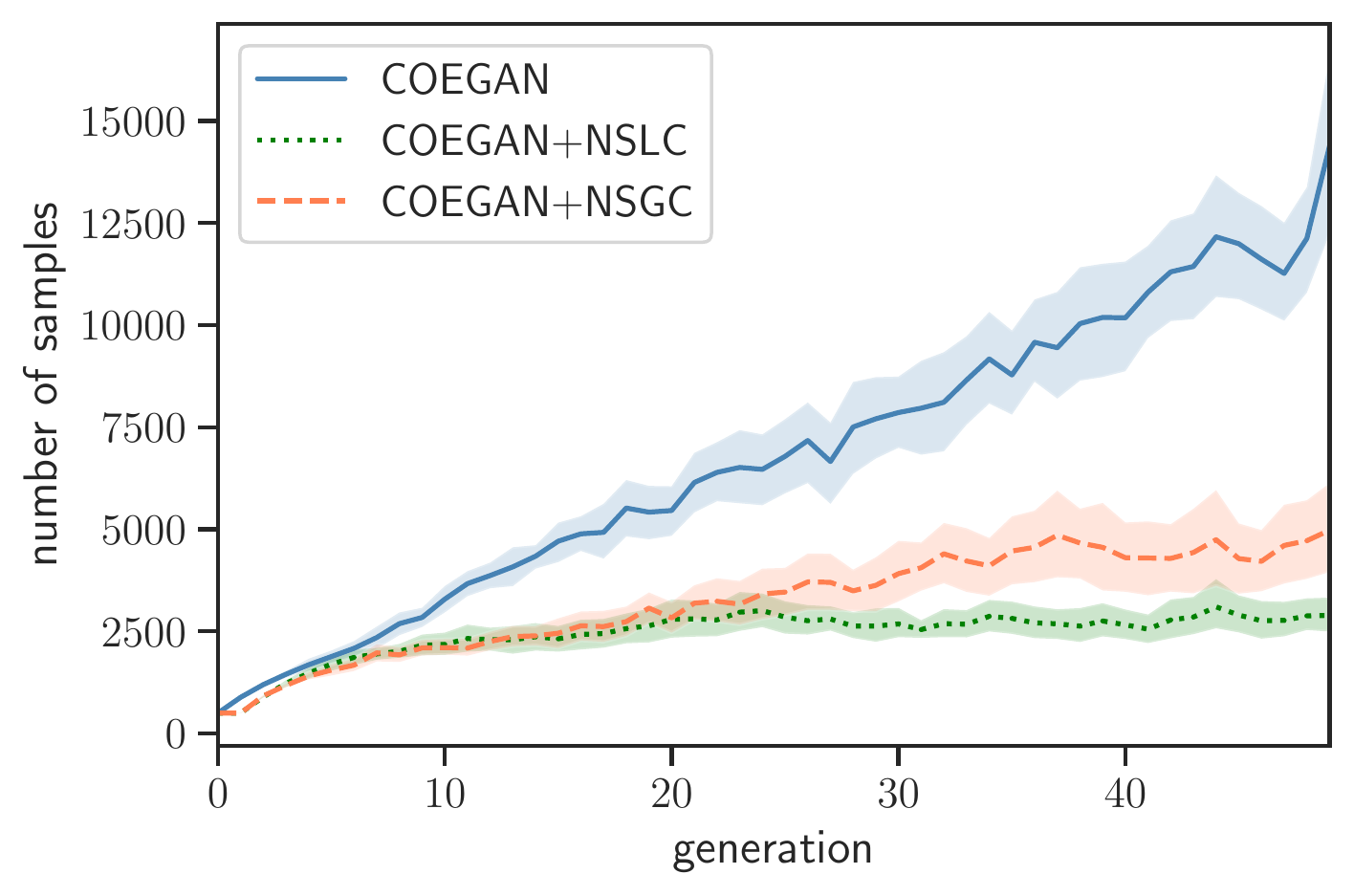}
	\caption{Average number of samples used to train all discriminators with MNIST.}
	\label{fig:mnist_genes_used_d}
\end{figure}

\begin{figure}[h]
	\includegraphics[width=0.4\textwidth]{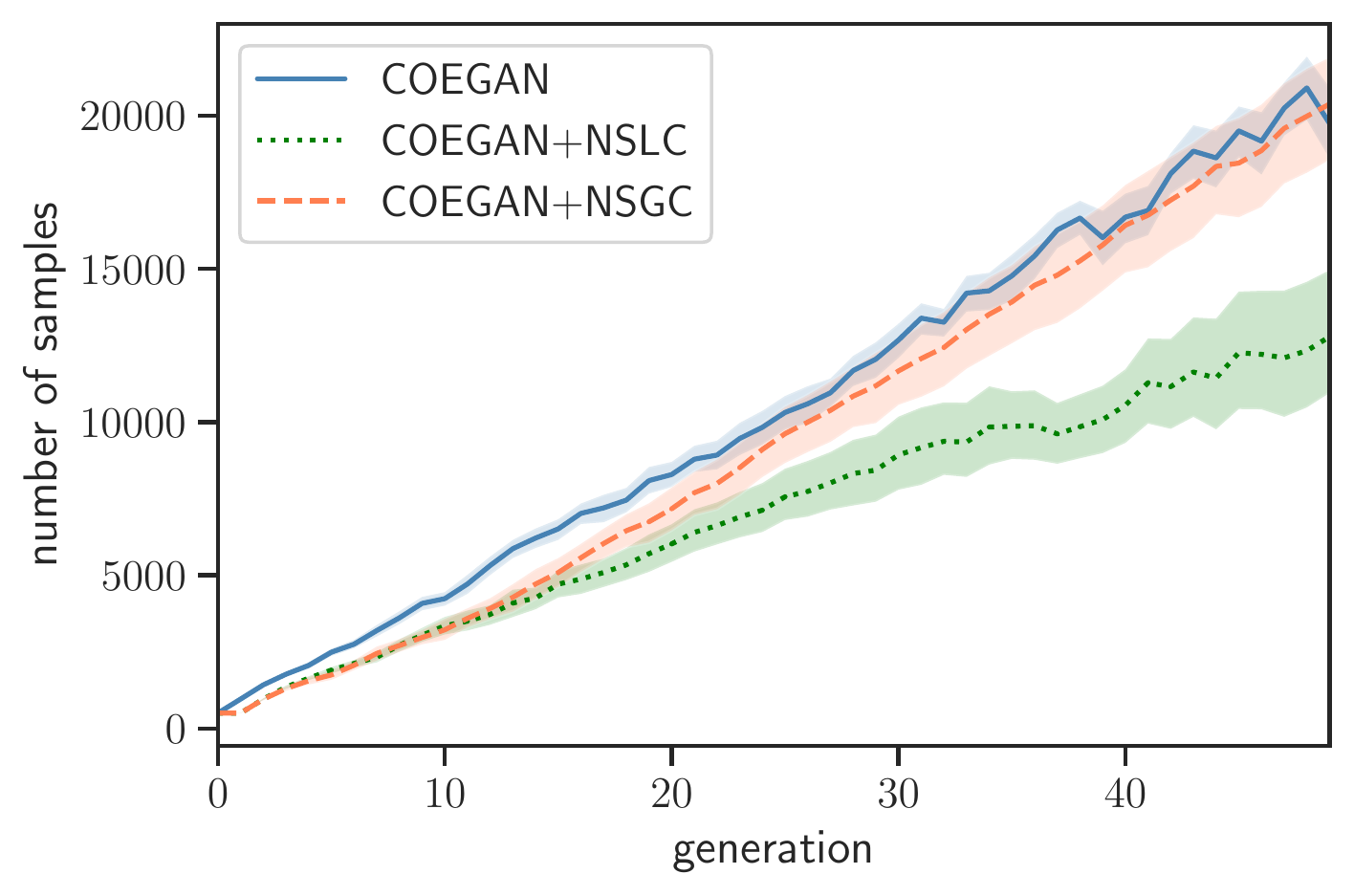}
	\caption{Average number of samples used to train all generators with MNIST.}
	\label{fig:mnist_genes_used_g}
\end{figure}

Figures \ref{fig:mnist_genes_used_d} and \ref{fig:mnist_genes_used_g} display the average number of samples used in the training process for all discriminators and generators in the population, respectively.
In these charts, we confirm the effect of the novelty strategy applied in our solution, which is more evident in Figure~\ref{fig:mnist_genes_used_d}.
These results evidence that newer individuals were more frequently selected through generations in the solutions based on the QD algorithm, resulting in fewer training samples directly seen by them~(new individuals can have new genes introduced by variation operators).
As expected, we also show that novelty is more present in the local competition solution when compared to the global competition version.

\begin{figure}[h]
	\includegraphics[width=0.4\textwidth]{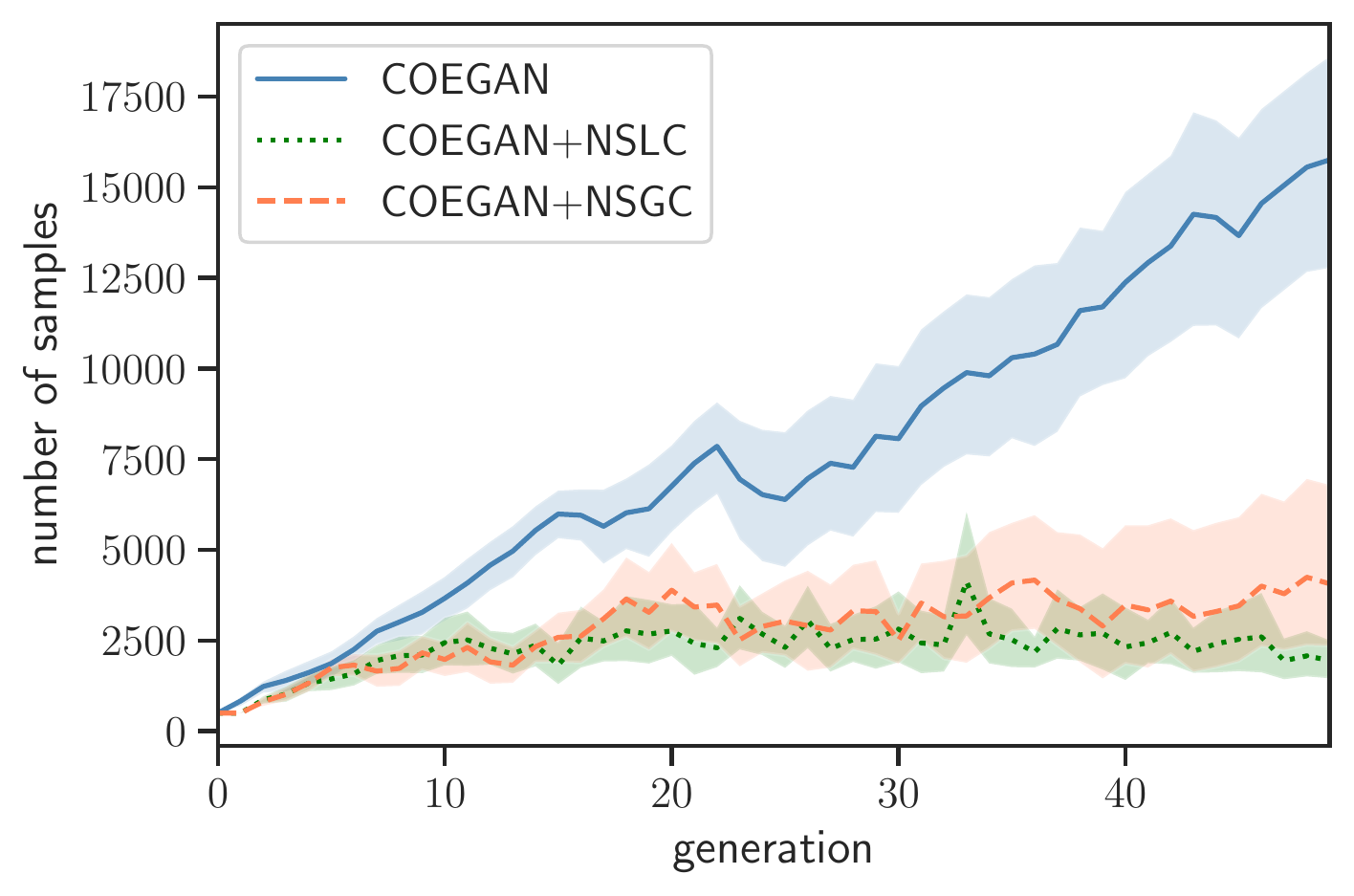}
	\caption{Average number of samples used to train best discriminators with MNIST.}
	\label{fig:mnist_genes_used_d_best}
\end{figure}

\begin{figure}[h]
	\includegraphics[width=0.4\textwidth]{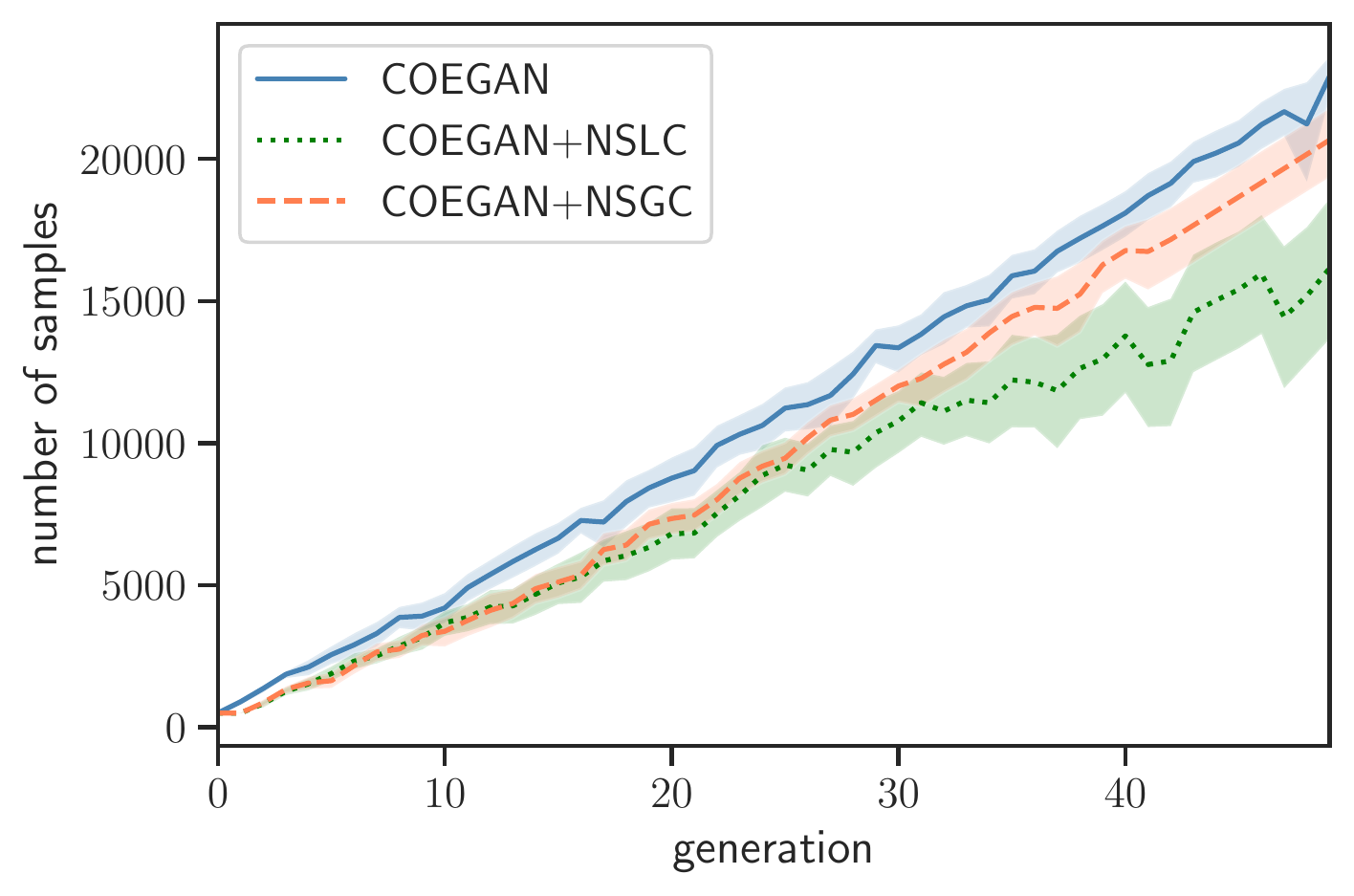}
	\caption{Average number of samples used to train best generators with MNIST.}
	\label{fig:mnist_genes_used_g_best}
\end{figure}

Figures~\ref{fig:mnist_genes_used_d_best} and \ref{fig:mnist_genes_used_g_best} provide additional support for this analysis, showing the number of samples used in training of the best individuals in the population of discriminators and generators, respectively.
The curves follow a similar behavior of Figures \ref{fig:mnist_genes_used_d} and \ref{fig:mnist_genes_used_g}, evidencing that best individuals at each generation also present more innovation in the solutions using COEGAN+NSLC.

A difference in the novelty effect is evident when comparing Figure~\ref{fig:mnist_genes_used_d} to Figure~\ref{fig:mnist_genes_used_g} and Figure~\ref{fig:mnist_genes_used_d_best} to Figure~\ref{fig:mnist_genes_used_g_best}.
The effect of innovation is more evident for discriminators (Figures~\ref{fig:mnist_genes_used_d} and \ref{fig:mnist_genes_used_d_best}) than in the results with generators (Figures~\ref{fig:mnist_genes_used_g} and \ref{fig:mnist_genes_used_g_best}).
We attribute this difference to the choice of fitness functions.
As concluded in~\cite{costa2019evaluating}, the FID score used in generators is a more reliable metric than the loss function used in discriminators.
This affects the quality criterion used in the NSGA-II optimization method, making the selection of better individuals more assertive.
However, further experiments are required to confirm this effect on innovation in the populations.

\begin{figure}[ht]
	\centering
	\begin{subfigure}[t]{0.23\textwidth}
		\includegraphics[width=\linewidth]{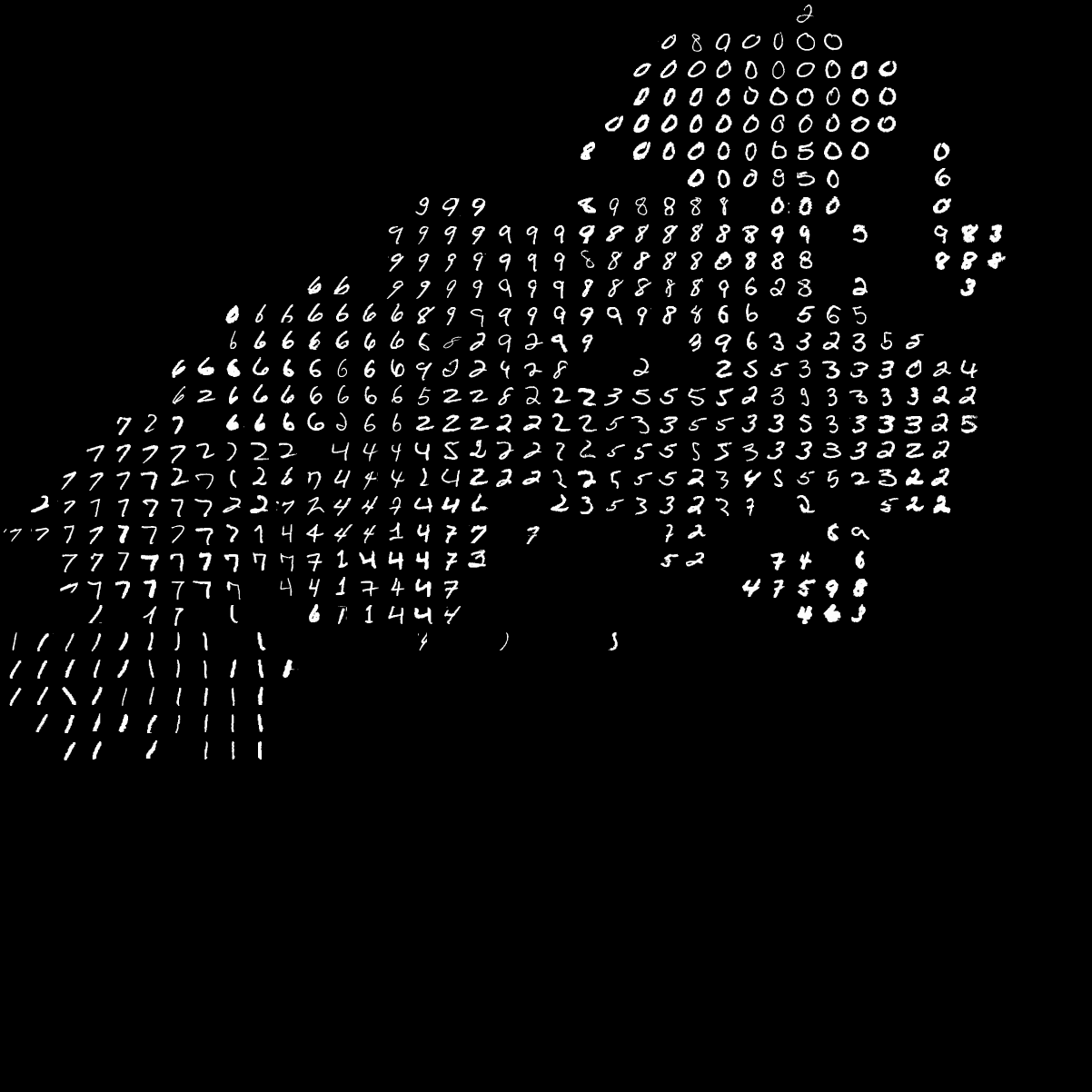}
		\caption{MNIST dataset\\$1121$ overlapped samples}
		\label{fig:mnist_tsne_dataset}
	\end{subfigure}\enskip%
	\begin{subfigure}[t]{0.23\textwidth}
		\includegraphics[width=\linewidth]{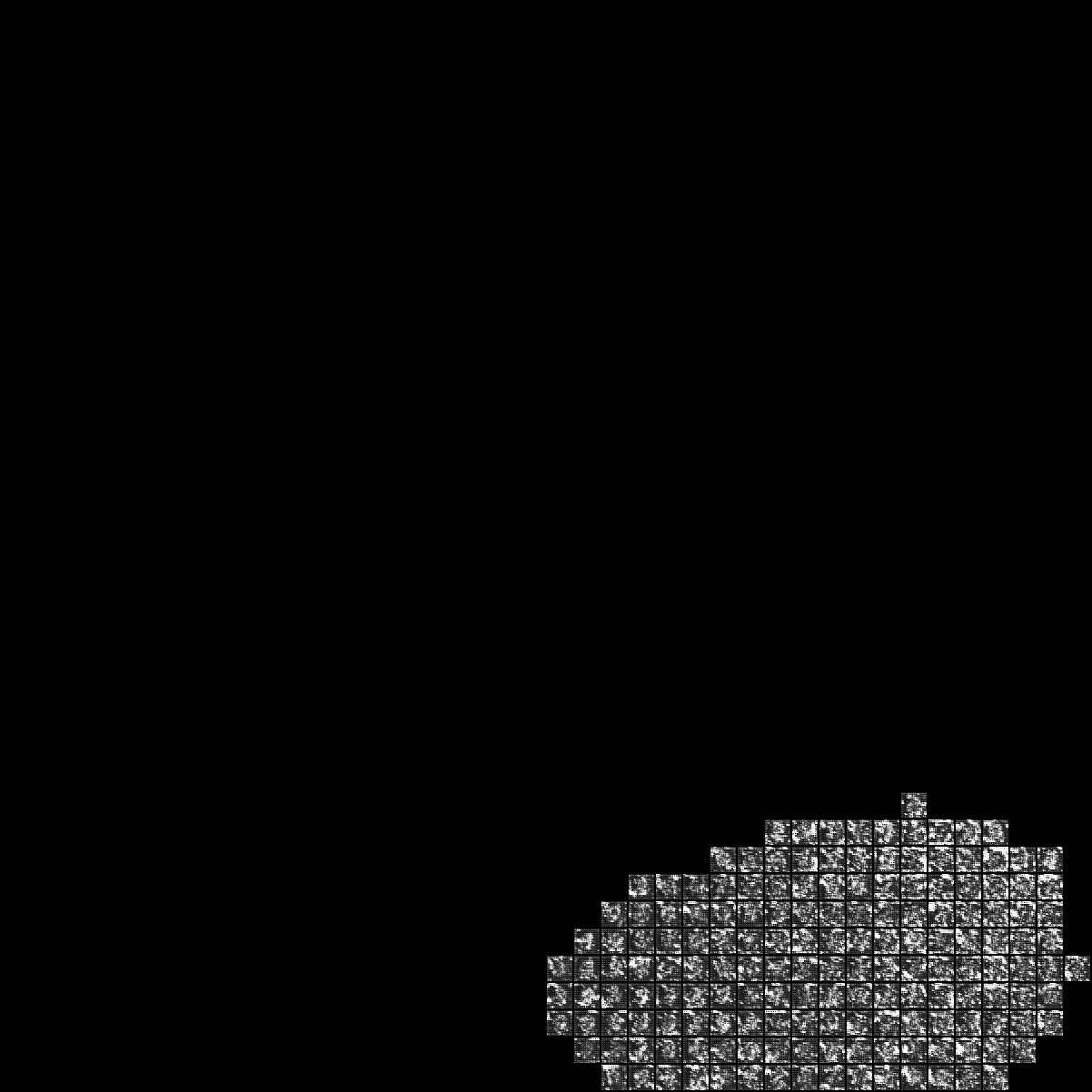}
		\caption{Generation 1\\$1436$ overlapped samples}
		\label{fig:mnist_tsne_0}
	\end{subfigure}\enskip%
	\begin{subfigure}[t]{0.23\textwidth}
		\includegraphics[width=\linewidth]{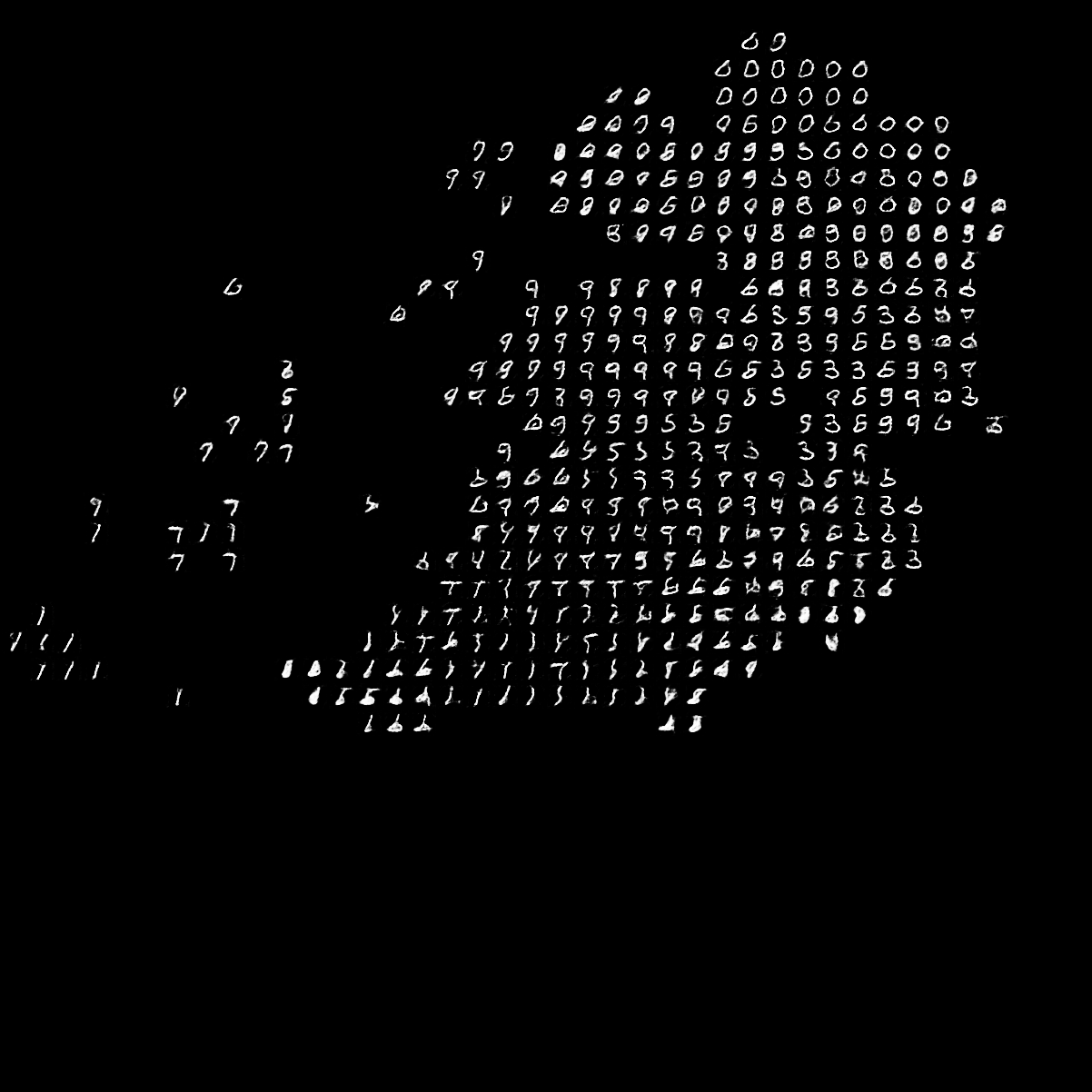}
		\caption{Generation 10\\$1189$ overlapped samples}
		\label{fig:mnist_tsne_09}
	\end{subfigure}\enskip%
	\begin{subfigure}[t]{0.23\textwidth}
		\includegraphics[width=\linewidth]{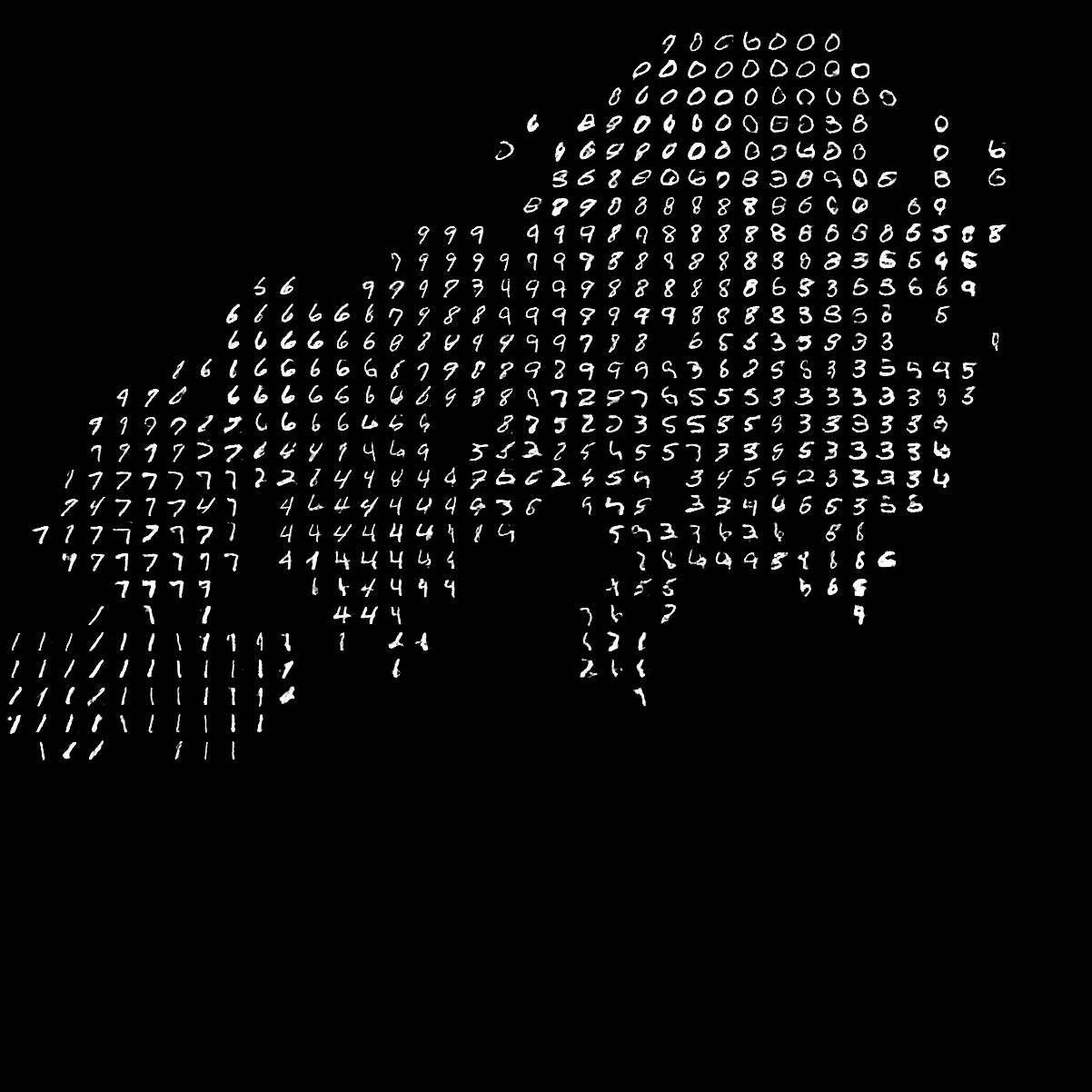}
		\caption{Generation 50\\$1077$ overlapped samples}
		\label{fig:mnist_tsne_49}
	\end{subfigure}
	\caption{Distribution of samples using t-SNE with the MNIST Dataset. We show samples \subref{fig:mnist_tsne_dataset} from the input dataset, the best generator at the \subref{fig:mnist_tsne_0} first generation, \subref{fig:mnist_tsne_09} after ten generations, and \subref{fig:mnist_tsne_49} at the end of training. We fed t-SNE with $1600$ samples from each scenario and used the results for positioning them into a two-dimensional space. The number of overlapped samples is displayed for each case.}
	\label{fig:mnist_tsne}
\end{figure}

To better study the quality and diversity achieved by our model, we present in Figure~\ref{fig:mnist_tsne} the distribution of samples produced by the best generator at different steps of the training process in one execution using the COEGAN+NSGC approach.
Samples are placed in a $40\times40$ grid, positioned according to t-SNE~\footnote{An expanded version of these images using a $120\times120$ grid is available at https://github.com/vfcosta/qd-coegan.}.
For this, $1600$ samples from each case were used in the t-SNE training and a discretization function is applied to place these samples into the two-dimensional space.
This method results in some overlapping samples, which indicates the level of diversity obtained by a model, i.e., fewer overlapping samples is evidence of better representation of the latent space.

Figure~\ref{fig:mnist_tsne_dataset} represents the distribution of the MNIST dataset, i.e., the variety of samples used in training.
We can see in Figure \ref{fig:mnist_tsne_0} that, at the initial stage, the samples are noisy and do not resemble images from MNIST, creating a high number of overlapping samples.
In generation $10$, represented by Figure \ref{fig:mnist_tsne_09}, the distribution of samples is more close to the presented in Figure~\ref{fig:mnist_tsne_dataset}, although we can still see some lack of quality and under-representation of some digits.
Figure \ref{fig:mnist_tsne_49} shows samples produced after the whole evolutionary process.
These samples have better quality and preserve diversity, resulting in $1077$ overlapping samples, even lower than the $1121$ overlapping samples presenting in the MNIST dataset.

\begin{figure}[htb]
	\includegraphics[width=0.4\textwidth]{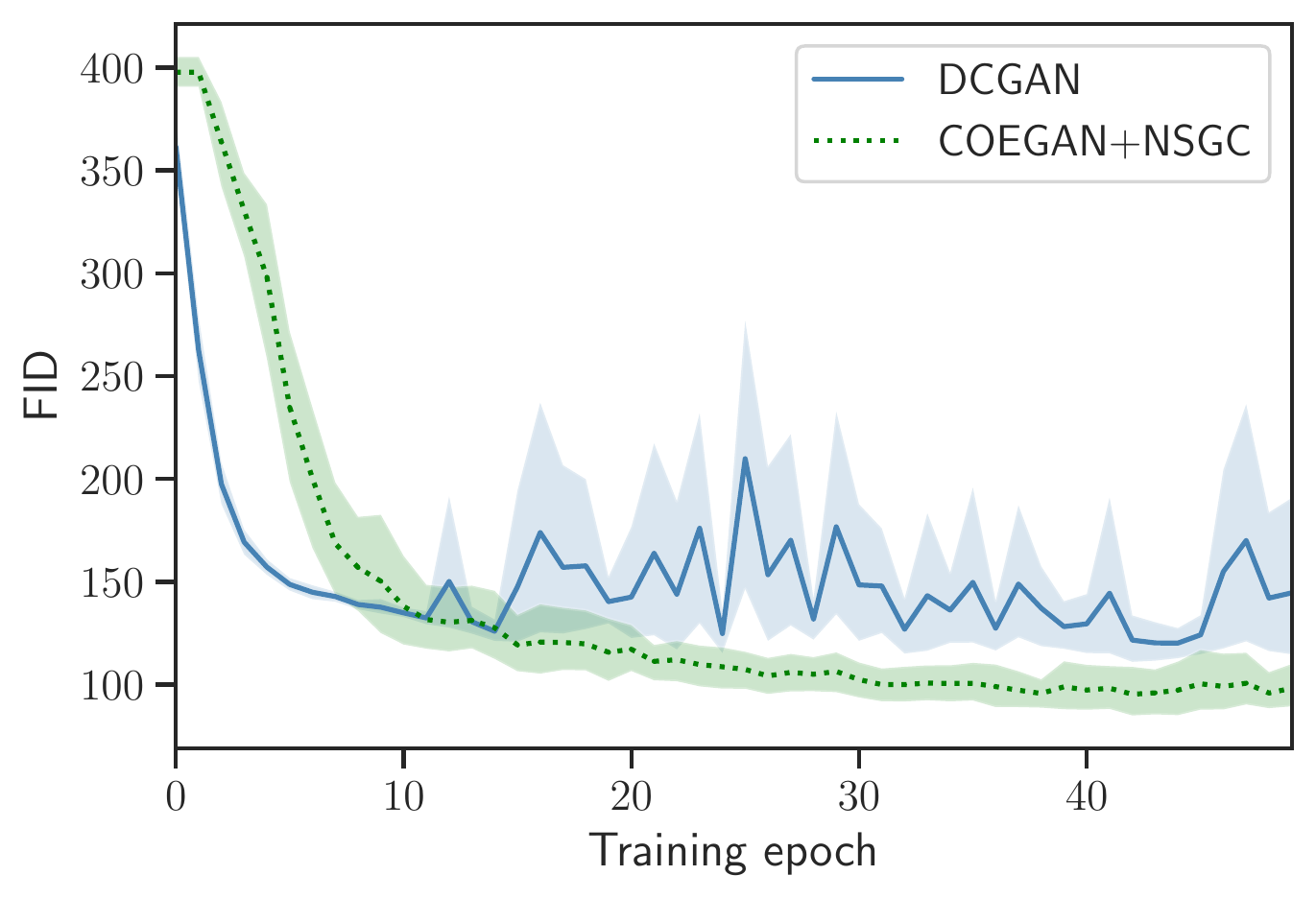}
	\caption{Best FID Score on the CelebA dataset.}
	\label{fig:celeba_fid_score_g_best}
\end{figure}

To assess the efficiency of our solution in complex datasets, we used COEGAN+NSGC, the best performing version of the proposed algorithm, to conduct experiments with CelebA~\cite{liu2015deep}.
For the sake of simplicity, we reduced the type of layers only to convolution and transpose convolution layers when adding a new gene, excluding the linear layer from the set of possibilities.
Besides, the activation functions were restricted to ReLU and Leaky ReLU in the mutation operators.
The populations of generators and discriminators contain $5$ individuals each.
The images from the CelebA dataset were rescaled to $64\times64$.
To handle images of bigger sizes, we increased the genome limit to $5$ and the number of batches per generation to $200$.
The remainder of the parameters is the same presented in Table~\ref{table:setup}.

We compare the results of our approach with a regular GAN that uses an architecture based on DCGAN~\cite{radford2015unsupervised}.
In the DCGAN-based experiments, the architecture of the generator and the discriminator is composed of four layers.
Previous experiments with neural networks using five layers were conducted but the results were more unstable, making the four-layers version more suitable for comparison with COEGAN+NSGC.
It is important to note that we ensure the DCGAN approach is trained with the same number of samples of an individual in COEGAN+NSGC.
Therefore, we define one training epoch as the training of DCGAN with $1000$ batches.
For COEGAN+NSGC, one training epoch is equivalent to one generation of the evolutionary algorithm.
As we use the \textit{all vs. all} pairing approach, each individual is also trained with $1000$ batches per generation ($5$ individuals times $200$ batches).

Figure~\ref{fig:celeba_fid_score_g_best} presents the progression of the best FID score when training with CelebA.
We can see a smooth progression of the FID in the COEGAN+NSGC approach, leading to a final result consistently better when compared to the DCGAN-based solution.
In DCGAN, FID varies during the training epoch, demonstrating spikes during the process mostly due to the occurrence of common stability issues on the GAN training, such as the mode collapse problem~\cite{brock2018large}.
This is evidence that our approach provides more stability on GAN training when compared to regular GANs with similar architectures.

\begin{figure}[htb]
	\centering
	\begin{subfigure}[t]{0.22\textwidth}
		\includegraphics[width=\textwidth]{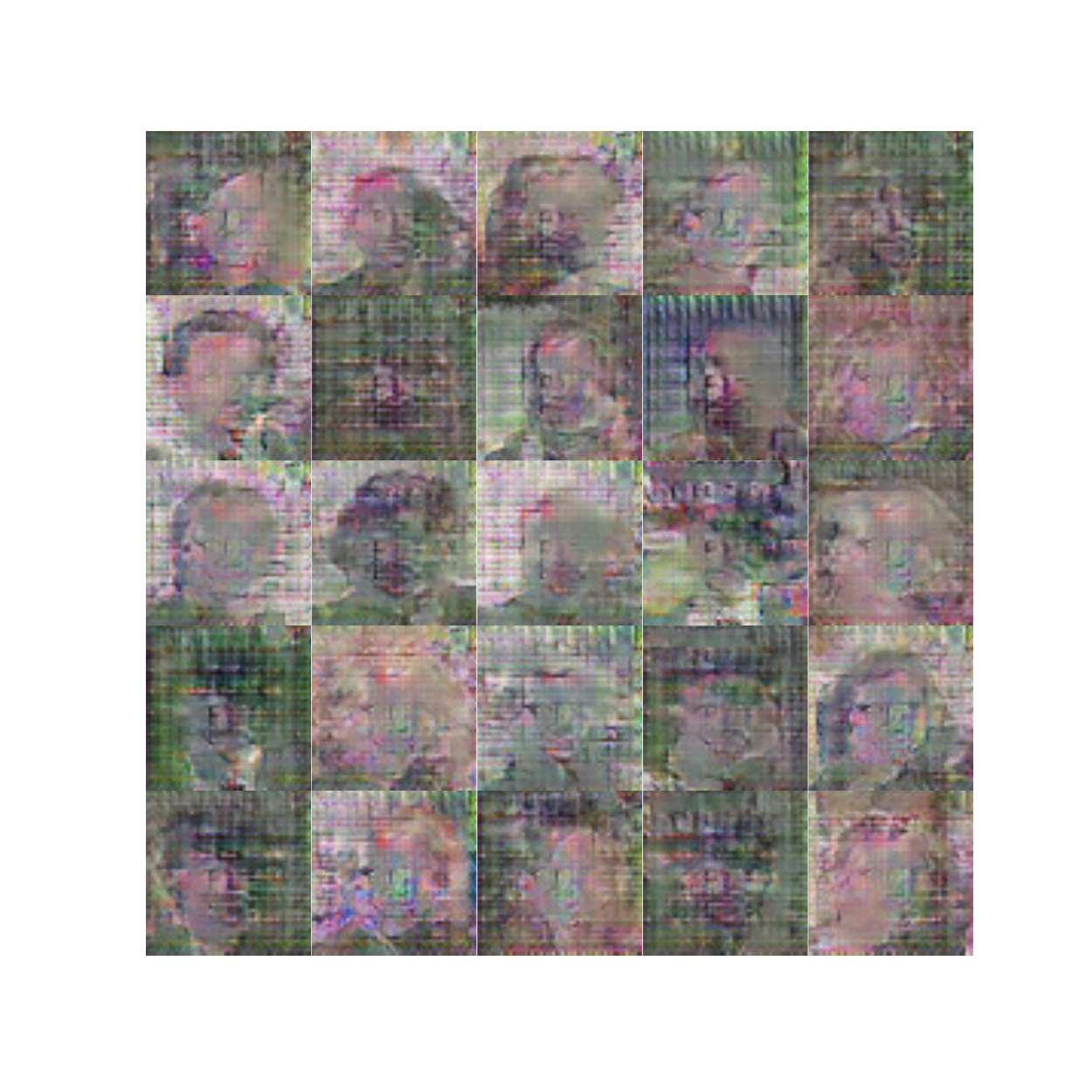}
		\caption{DCGAN}
		\label{fig:celeba_samples_collapse}
	\end{subfigure}\enskip%
	\begin{subfigure}[t]{0.22\textwidth}
		\includegraphics[width=\textwidth]{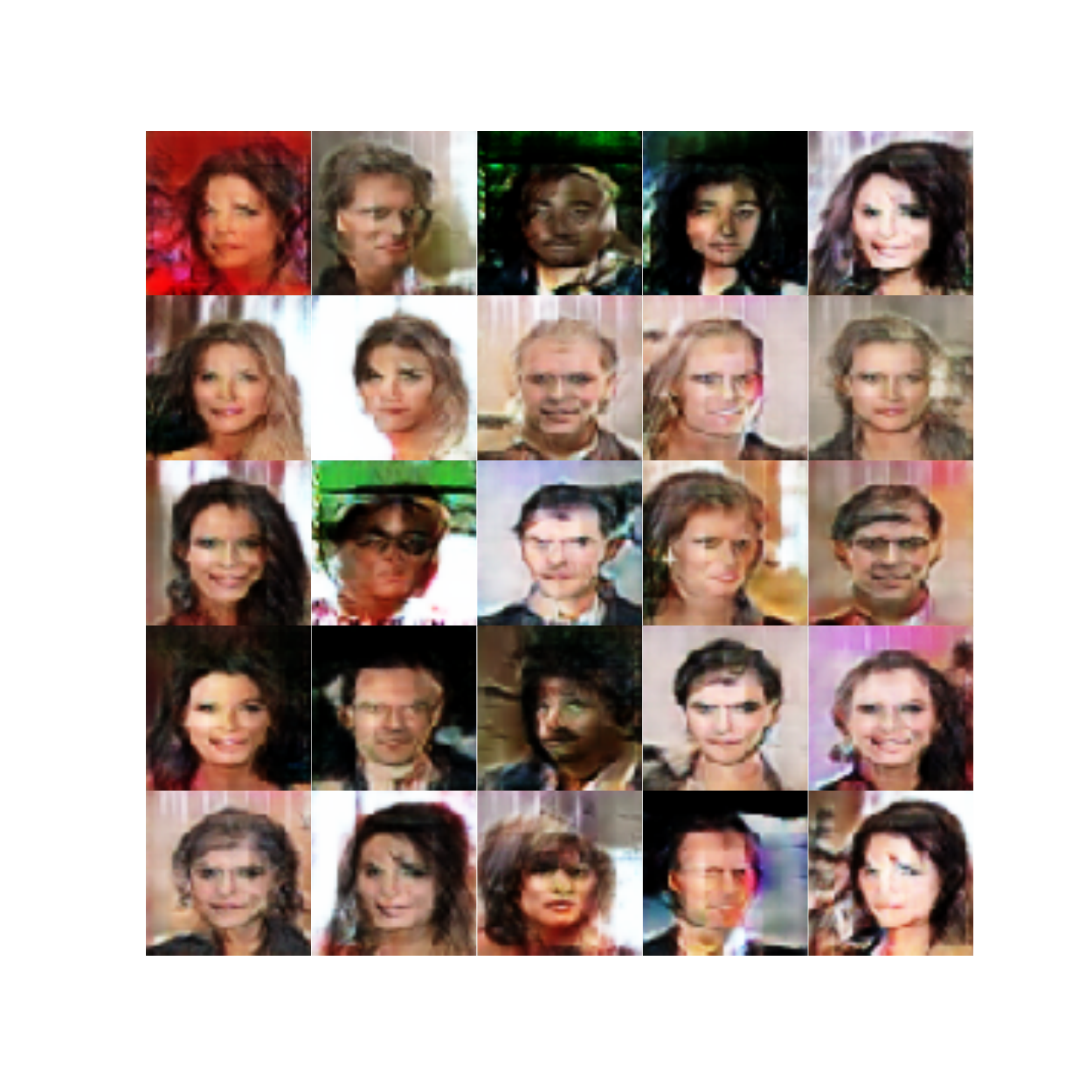}
		\caption{COEGAN+NSGC}
		\label{fig:celeba_samples_nsgc}
	\end{subfigure}
	\caption{Samples created by \subref{fig:celeba_samples_collapse} DCGAN after collapsing in final training epoch and by \subref{fig:celeba_samples_nsgc} COEGAN+NSGC after training.}
	\label{fig:celeba_samples}
\end{figure}

Figure~\ref{fig:celeba_samples} shows samples created by both approaches when trained with the CelebA dataset.
Figure~\ref{fig:celeba_samples_collapse} displays samples created by the DCGAN approach after the final epoch when issues were observed in training.
This is an example of the resulting effect of a spike that occurred in the DCGAN training.
In Figure~\ref{fig:celeba_samples_nsgc} we can see samples produced after training with COEGAN+NSGC.
Although spikes are not present, the quality of samples produced by COEGAN+NSGC is not perfect, being worse than state-of-the-art GANs trained with CelebA~\cite{zhang2018stackgan++}.
Further experiments should be executed to assess the capability of COEGAN+NSGC to achieve better results concerning the FID score in larger experimental setups.

\section{Conclusions} \label{sec:conclusions}
In this paper, we have investigated the application of a quality diversity algorithm to train and evolve Generative Adversarial Networks (GANs).
GANs are capable of producing strong generative models, but the training procedure is hard and stability issues frequently affect the results.
Furthermore, the hand-design of efficient models is a time-consuming task, worsen by the inconstancy of the GAN training.

Evolutionary algorithms were recently proposed to improve the training of GANs and to provide the discovery of efficient generative models.
COEGAN uses coevolution with an evolutionary algorithm inspired by NEAT to train GANs.
However, the lack of diversity and premature optimization leave room for improvement of the solution.

We propose in this paper the extension of COEGAN to use a Quality Diversity (QD) algorithm in order to improve the exploration of the search space.
Therefore, we design a new evolutionary algorithm that combines COEGAN with the approach used in the Novelty Search with Local Competition (NSLC) algorithm.

The experimental results show that the use of QD to guide the evolution of GANs improved the diversity in the population, leading to the discovery of better models.
Furthermore, experiments with a version of the algorithm using global competition evidence that we can consistently outperform the previous results of COEGAN in the MNIST dataset.
Experiments with the CelebA dataset indicate that our proposal provides a more stable training when compared to a regular GAN based on the DCGAN architecture, avoiding problems such as mode collapse and vanishing gradient.

As further works, we pretend to explore our quality diversity approach and extend the experimental setup to increase the capability to discover better models when training in complex datasets.
Besides, new architectural components recently proposed for GANs can be incorporated into the model to enhance the population of individuals represented by our solution.

\section*{Acknowledgments}\label{sec:acknowledgments}
This work is funded by national funds through the FCT - Foundation for Science and Technology, I.P., within the scope of the project CISUC - UID/CEC/00326/2020 and by European Social Fund, through the Regional Operational Program Centro 2020.

\bibliographystyle{ACM-Reference-Format}
\bibliography{references}

\end{document}